\documentclass{article}

\PassOptionsToPackage{numbers,sort&compress}{natbib}

\usepackage[preprint]{neurips_2025}

\workshoptitle{New Perspective in Graph Machine Learning (NPGML)}

\usepackage{multirow} 
\usepackage{graphicx}
\usepackage{amsmath}
\usepackage[T1]{fontenc}    
\usepackage{hyperref}       
\usepackage{url}            
\usepackage{booktabs}       
\usepackage{amsfonts}       
\usepackage{nicefrac}       
\usepackage{microtype}      
\usepackage{xcolor}         
\usepackage{float}
\title{Predicting Microbial Interactions Using Graph Neural Networks}

\author{%
  \textbf{Elham Gholamzadeh} \\
 Max Planck Institute for Human Cognitive and Brain Sciences \\
  Center for Scalable Data Analytics and Artificial Intelligence (ScaDS.AI) \\
  Dresden/Leipzig, Germany \\
  \texttt{gholamzade@cbs.mpg.de}
  \and
  \textbf{Kajal Singla} \\
  Max Planck Institute for Human Cognitive and Brain Sciences \\
  Center for Scalable Data Analytics and Artificial Intelligence (ScaDS.AI) \\
  Dresden/Leipzig, Germany \\
  \texttt{singla@cbs.mpg.de}
  \and
  \textbf{Nico Scherf} \\
  Max Planck Institute for Human Cognitive and Brain Sciences \\
  Center for Scalable Data Analytics and Artificial Intelligence (ScaDS.AI) \\
  Dresden/Leipzig, Germany \\
  \texttt{nscherf@cbs.mpg.de}
}

\begin{document}

\maketitle

\begin{abstract}
Predicting interspecies interactions is a key challenge in microbial ecology, as these
interactions are critical to determining the structure and activity of microbial communities.
In this work, we used data on monoculture growth capabilities, interactions with other species, and phylogeny to predict a negative or positive effect of interactions. More precisely, we used one of the largest available pairwise interaction datasets to train our models, comprising over 7,500 interactions between 20 species from two taxonomic groups co-cultured under 40 distinct carbon conditions, with a primary focus on the work of Nestor et al.\cite{nestor2023interactions}. In this work, we propose Graph Neural Networks (GNNs) as a powerful classifier to predict the direction of the effect. We construct edge-graphs of pairwise microbial interactions in order to leverage shared information across individual co-culture experiments, and use GNNs  to predict  modes of interaction. Our model can not only predict binary interactions (positive/negative) but also classify more complex interaction types such as mutualism, competition, and parasitism. Our initial results were encouraging, achieving an F1-score of 80.44\%. 
This significantly outperforms comparable methods in the literature, 
including conventional Extreme Gradient Boosting (XGBoost) models, 
which reported an F1-score of 72.76\%.

\end{abstract}

\section{Introduction}

Understanding microbial species' functions and interactions for developing structured microbial communities and processes is experimentally challenging and remains a fundamental problem in biology and biotechnological applications \cite{nadell2016spatial,wondraczek2019artificial}. Gaining a deeper understanding of these interactions can significantly enhance our capacity to manage and manipulate ecological communities, offering broad applications in environmental conservation, crop health, and human health  \cite{kavino2018vitro,wallace2013human}. For several decades, microbial communities have been used for the clean-up of contaminated water, soil, and air  and the capture of renewable resources such as energy (biogas and H2), chemicals (short-chain acids), and water \cite{tsoi2019emerging,eng2019microbial}. Studying microbial interactions exhaustively through pairwise co-culture experiments becomes increasingly impractical with increasing number of species, since the possible combinations rise exponentially. This limits our ability to experimentally search for key positive interactions that drive community function. Designing such microbial systems for bioengineering applications is further complicated by the complexity introduced by spatial organization and temporal dynamics. This suggests that much can be gained from reducing the parameter space of the system, e.g., by supervised machine learning methods. Supervised machine learning algorithms \cite{goodswen2021machine, reel2021using}, including support vector machines \cite{yang2021screening}, neural networks \cite{rampelli2021g2s}, random forests \cite{dicker2021sputum}, and graph neural networks \cite{pan2024microbial} to predict more complex phenotypes of microbial interactions \cite{cesario2021personalized, marcos2021applications, hughes2021high}, a task made more feasible by the rise of high-throughput techniques for generating large biological data sets.
Predicting interspecific interactions is critical for both understanding and engineering microbial communities, as the collective behavior of these systems is shaped by intricate relationships between species. These interactions can be negative where one species inhibits another through nutrient competition or the production of antimicrobial compounds \cite{ghoul2016ecology} or positive, where one species enhances the growth of another by increasing nutrient availability or creating new ecological niches \cite{bruno2003inclusion}. Accurate prediction of these relationships becomes especially important in species-rich environments or communities composed of metabolically fastidious organisms. Traditionally, such predictions have relied on genome-scale metabolic and constraint-based modeling approaches \cite{shoaie2013understanding, lara2021using}, which use genomic information to estimate interactions by assessing overlaps and complementarities in metabolic capabilities \cite{gu2019current, kim2012recent}.

Building on the work of Nestor et al.\cite{nestor2023interactions}, we leverage one of the largest data sets of experimentally validated microbial interactions to evaluate how well graph neural networks can predict the sign of microbial interactions (positive/negative) compared to other standard machine learning models. Similar to \cite{nestor2023interactions} we use the species’ phylogeny and their monoculture yield, which belong to each of the 40 carbon environments, as features to build our models. All pairwise interactions between 20 distinct soil bacteria from two taxonomic groups that were cultivated in 40 different environments \cite{nestor2023interactions}, each containing either a single carbon source or a combination of all carbon sources, are included in this data set.

\begin{figure}[ht]
  \centering
  \includegraphics[width=0.61\textwidth]{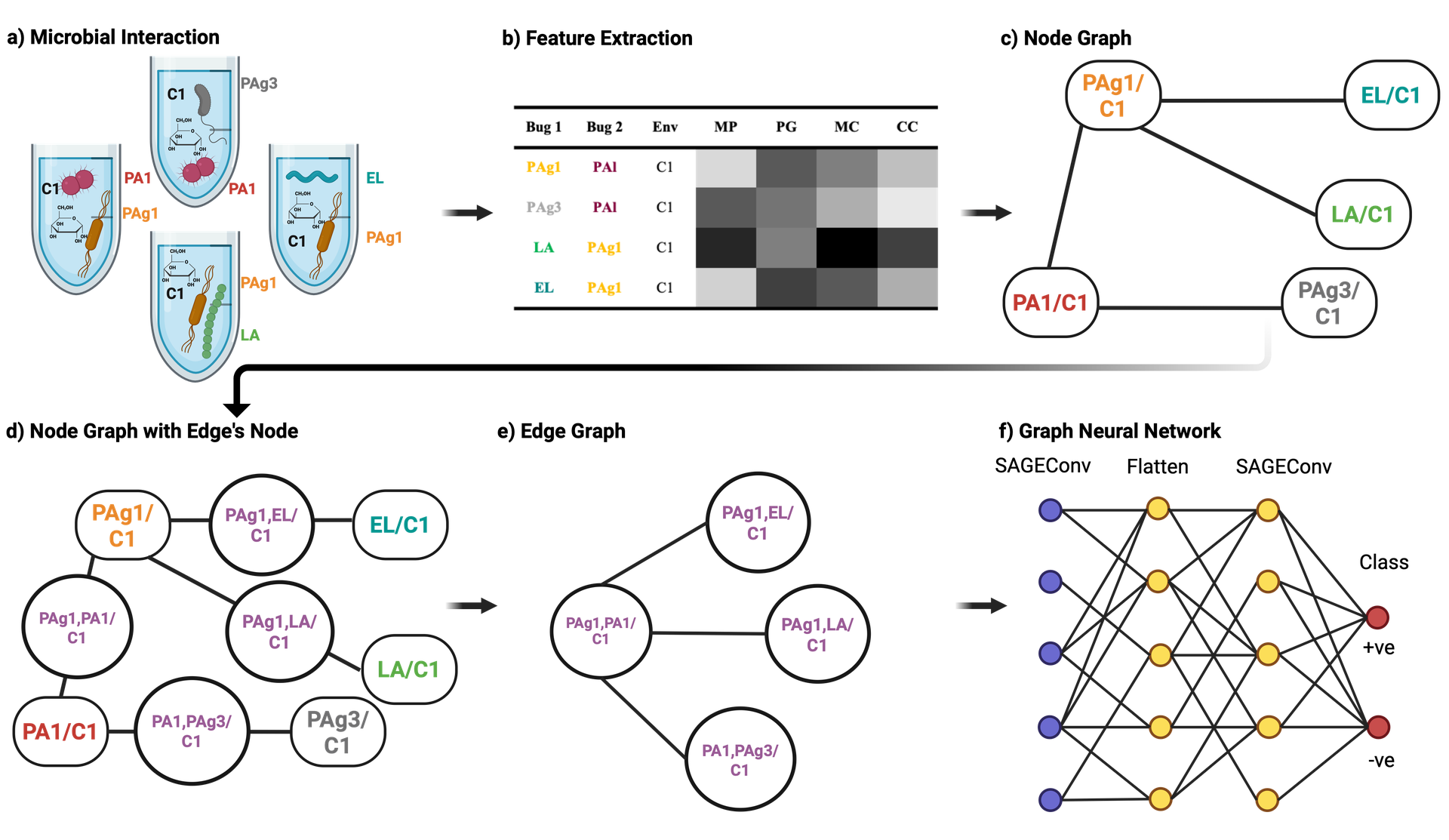}
  \caption{Illustration of leveraging graph neural networks to predict microbial interactions. This approach enables accurate prediction of both one-way and two-way interactions. a) Experimental measurements of microbial interaction; b) Feature creation;   c) Construct graph based on microbe/condition (nodes) and interactions measured by experiment (edges); d) Convert of the graph into a dual edge graph; e) Edge graph representation; f) Prediction using a graph neural network.}
  \label{fig:gnn_interaction}
\end{figure}

We propose using GNNs as a data-driven model of microbial communities and interactions (Figure.\ref{fig:gnn_interaction}) as GNNs are able to leverage information across separate (but related) experimental conditions. We build a graph that connects co-culture experiments through shared species or experimental conditions, and use this structure to train Graph Neural Networks (GNNs) to predict interactions between species in (experimentally) unknown combinations. In our graph structure, nodes represent a unique combination of species and condition, and edges capture the interactions between two given species under the same experimental condition. To directly encode relational information across experiments, we construct an edge graph, where each interaction (edge) in the original graph is transformed into a new node, now representing a combination of two species in a specific experimental condition. Two nodes in the edge graph are now connected if their corresponding experiments share a common species (and condition).

.

\section{Graph Neural Networks}

\subsection{Graph Neural Networks }

The interactions among microbes inherently form a graph structure, enabling their representation as a network of connections. We modeled these interactions using a graph neural network (GNN) \cite{wu2020comprehensive}.
Graph Neural Networks (GNNs) extend traditional neural networks to work directly with graph-structured data, where a graph is formally represented as \( G = (V, E) \), where \( V \) is the set of vertices (nodes) and \( E \) is the set of edges. The neighborhood of a node \( v \in V \) is defined as \( N(v) = \{ u \mid \{ u, v \} \in E \} \), and each node \( i \) is associated with an attribute \( x_i \) for \( i \in [1, |V|] \).
GNNs operate iteratively through message passing steps, where node attributes are updated based on interactions with their neighbors. At each step, the update process is as follows:
For each node \( i \), the features of its neighbors \( j \in N(i) \) are retrieved. The node features \( x_i \) are updated by aggregating information from its neighbors:

\begin{equation}\label{eq:edge}
e(i, j) = g_e(\mathbf{x}_i, \mathbf{x}_j)
\end{equation}

\begin{equation}\label{aggr}
\mathbf{x}'_i = g_v\left(\mathbf{x}_i, \mathrm{aggr}_{j \in \mathcal{N}(i)} \left( e(i, j) \right)\right)
\end{equation}

where  \( g_v \) and \( g_e \) are the node and edge update functions that represent an aggregation operation. Here, mean  is used as the aggregation function over the features of neighboring nodes.

\subsection{Edge-Graph Construction}

As illustrated in Figure~\ref{fig:gnn_interaction}, our framework construct the relationships between different species and their available features using an edge-graph representation. This approach focuses on relationships between experiments (edges) rather than solely between individual species (nodes), allowing us to explicitly encode dependencies between experimentally measured species interactions. The edge-graph \( L(G) \) provides a structural representation of edge adjacencies, making it particularly suitable for edge-centric tasks in Graph Neural Networks (GNNs).

We begin with the species interaction graph \( G = (V, E) \), where:
\begin{itemize}
    \item \( V \) denotes the set of species recorded under a shared conditions;
    \item \( E \) edges denote a shared species and condition between experiments.
\end{itemize}

The transformation from \( G \) to its edge-graph \( L(G) = (V', E') \) is defined formally as follows:
\begin{itemize}
    \item \textbf{Node mapping:} For each interaction \( e_k = (u, v) \in E \), create a node \( v'_k \in V' \) in \( L(G) \). Each \( v'_k \) represents a single species interaction.
    \item \textbf{Adjacency mapping:} Two nodes \( v'_i, v'_j \in V' \) are connected by an edge in \( E' \) if and only if their corresponding edges in \( G \) share at least one endpoint species and experimental condition.
    
    \item \textbf{Feature mapping:} Any attributes originally associated with the interaction \( e_k \) (e.g., monogrowth characteristics, phylogenetic tree–based distances, etc.) are assigned to the corresponding node \( v'_k \) in \( L(G) \).

\end{itemize}

This construction turns interaction–interaction dependencies into direct connections in \( L(G) \). It allows message passing to operate over interactions themselves, capturing higher‑order ecological relationships.

\subsection{Proposed Model}

To classify microbial interactions, we implemented a two-layer GraphSAGE model using the Deep Graph Library (DGL) \cite{hamilton2017inductive}. Each node in the graph represents a microbial species, and edges capture potential interactions. The task is to assign a binary label \( y_i \in \{+, -\} \) to each node \( v_i \).

The two layers of the model apply GraphSAGE convolutions with mean aggregation, allowing each node to iteratively incorporate feature information from its local neighborhood. In GraphSAGE, the edge function \( e(i, j) \) outputs the attributes of a neighboring node \( j \in \mathcal{N}(i) \):
\begin{equation}
g_e(\mathbf{x}_i, \mathbf{x}_j) = \mathbf{x}_j,
\label{eq:edge_function}
\end{equation}
and the node update function \( g_v \) is defined as
\begin{equation}
\mathbf{x}'_i = W_1 \mathbf{x}_i + W_2 \cdot \mathrm{mean}_{j \in \mathcal{N}(i)} \mathbf{x}_j,
\label{eq:node_update}
\end{equation}
where \( W_1 \) and \( W_2 \) are learnable weight matrices.

The ReLU activation function is applied after first layer to introduce non linearity. No explicit edge features are used in the model learning is driven entirely by node attributes and graph connectivity.

The input feature size is 13 (see Table~\ref{tab:features}), and ReLU is used as the non-linear activation function:
\begin{equation}
\text{ReLU}(x) = \max(0, x)
\end{equation}

We optimize the model using cross-entropy loss:
\begin{equation}
\mathcal{L}(\hat{y}, y) = -\sum_{i=1}^{N} y_i \log(\hat{y}_i),
\end{equation}
where \( \hat{y} \) is the predicted label and \( y \) is the ground-truth.

This configuration allows for inductive learning and generalization to unseen nodes or graphs, making it well-suited for microbial community analysis.

\section{MATERIALS }\label{444}
\subsection{Dataset}
The dataset comprises over 7{,}500 pairwise interactions among 20 species from two taxonomic groups across 40 carbon environments. It is primarily based on the works of Nestor \textit{et al.}~\cite{nestor2023interactions} and Kehe \textit{et al.}~\cite{kehe2019massively}, which employ the kChip platform a high-throughput, nanodroplet-based system for combinatorial screening to evaluate species growth in monoculture and coculture settings.
\subsubsection{Features creation}
By   primarily focusing on the \cite{nestor2023interactions}, to enhance the predictive power of our model, monoculture growth yields for each species and supplementary features, such as the species' phylogenetic relationships and their metabolic profiles across the carbon environments, are incorporated. The phylogenetic data was summarized using principal component analysis (PCA), where the first two principal components captured 95\% of the variance. Similarly, the metabolic profiles were represented using the first four principal components, which accounted for 90\% of the variation.
These reduced-feature phylogenetic data and metabolic profiles were used to construct a feature set that effectively captures both the evolutionary and metabolic similarities between species. Additionally, the Euclidean distance between the monoculture growth profiles of interacting species was computed to further describe their metabolic relationship.
This feature set, which is presented in (Table.\ref{tab:features}  in \ref{appendix:features} Appendix),  allowed  us to model and predict the sign effect of interactions, enabling better understanding of species relationships in diverse carbon environments.

\subsection {Model implementation}
We evaluated our method using standard classification metrics, including accuracy (Acc), sensitivity (Sen or Recall), precision (Prec), and F1 score (equations are provided in Appendix~\ref{appendix:metrics})
We implemented the proposed solution in Python (version~3) using the GraphSAGE model from the PyTorch framework~\cite{paszke2019pytorch}. During training, we employed the Adam optimizer~\cite{kingma2014adam} to perform first-order gradient-based optimization on the stochastic objective function, with a learning rate set to \(1 \times 10^{-2}\). The dataset was split into 80\,\% for training and 20\,\% for testing. The performance of the GraphSAGE model was evaluated over 300 training epochs. All experiments were conducted on the RAVEN HPC system, equipped with Intel Xeon IceLake-SP processors and NVIDIA A100 GPU nodes interconnected via NVLink.

\section{Results}
\begin{figure}[ht]
  \centering
  \includegraphics[width=0.9
  \textwidth]{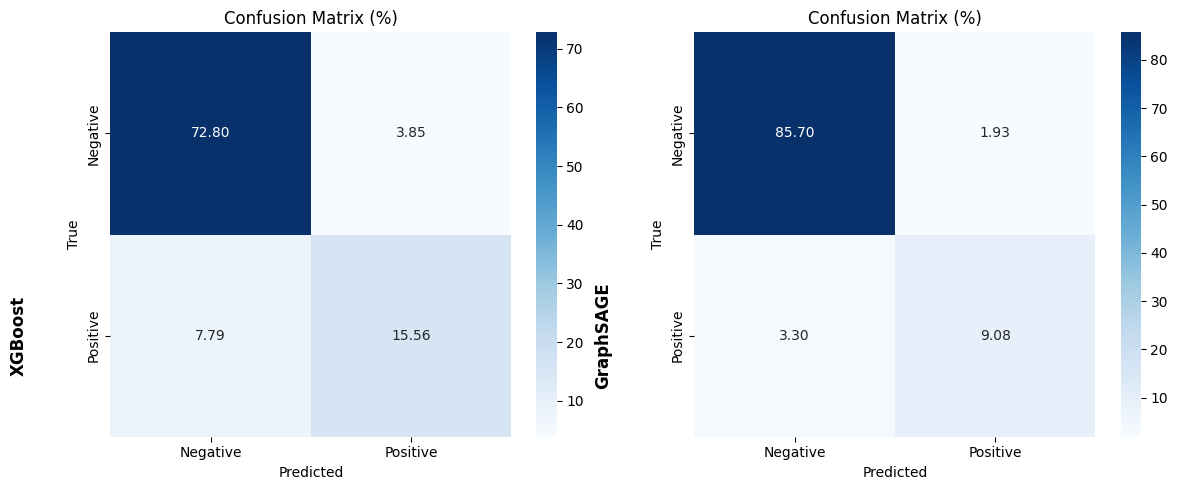
}
  \caption{Comparison of the confusion matrices for XGBoost and GraphSAGE models of one-way prediction.}\label{confution}
\end{figure}

 We compare our method with the standard machine learning algorithms used in \cite{nestor2023interactions} for predicting the sign of one- and two-way growth yield effects. Specifically, we evaluated the predictive performance of our approach against two baselines: k-Nearest Neighbors (kNN)  and Extreme Gradient Boosting (XGBoost). The Graph Neural Network achieved significantly higher accuracy in predicting the signs of both one- and two-way effects.
The results  for one-way prediction of different models are reported in Table.\ref{tab:model_metrics}. GraphSAGE learns these node embeddings in an end-to-end manner, optimizing them specifically for the classification objective and explicitly models the connection across experiments.
Figure.\ref{confution} shows the confusion matrix of the GraphSAGE model and XGBoost for one-sign prediction. Our results show a class imabalance as positive interactions appear to be rare in both models for the one-way interaction sign prediction. Thus, the F1-score ( harmonic mean of precision and recall) is a more informative metrics for an imbalanced class distribution. Our edge-graph-based models demonstrated strong performance, achieving an  F1-score of 80.44\%, which outperformed XGBoost (72.76\%) and k-NN (66.13\%).

\begin{table}[ht]
\centering
\caption{Model performance metrics of one-way prediction on test dataset.}
\label{tab:model_metrics}
\begin{tabular}{lcccc}
\toprule
\textbf{Model} & \textbf{Accuracy} & \textbf{Sensitivity} & \textbf{Precision} & \textbf{F1-Score} \\
\midrule
GraphSAGE   &\textbf{ 0.9528} & 0.7392 & \textbf{0.8822} & \textbf{0.8044} \\
K-nearest neighbors  \cite{nestor2023interactions}  & 0.8593 & 0.6613 & 0.6613 & 0.6613  \\

XGBoost \cite{nestor2023interactions} & 0.8835 & \textbf{0.8017} & 0.7276 & 0.7276 \\
\bottomrule
\end{tabular}
\end{table}


We further implemented the GraphSAGE framework to predict two-way interactions between species: mutualism (+,+), parasitism (+,-), and competition (-,-). This approach leverages the structural properties of ecological networks to model complex interspecies relationships. By integrating both direct and indirect effects of species on one another, the GNN facilitates accurate predictions of interaction types.
 
\begin{figure}[ht]
\centering
\includegraphics[scale=0.38]{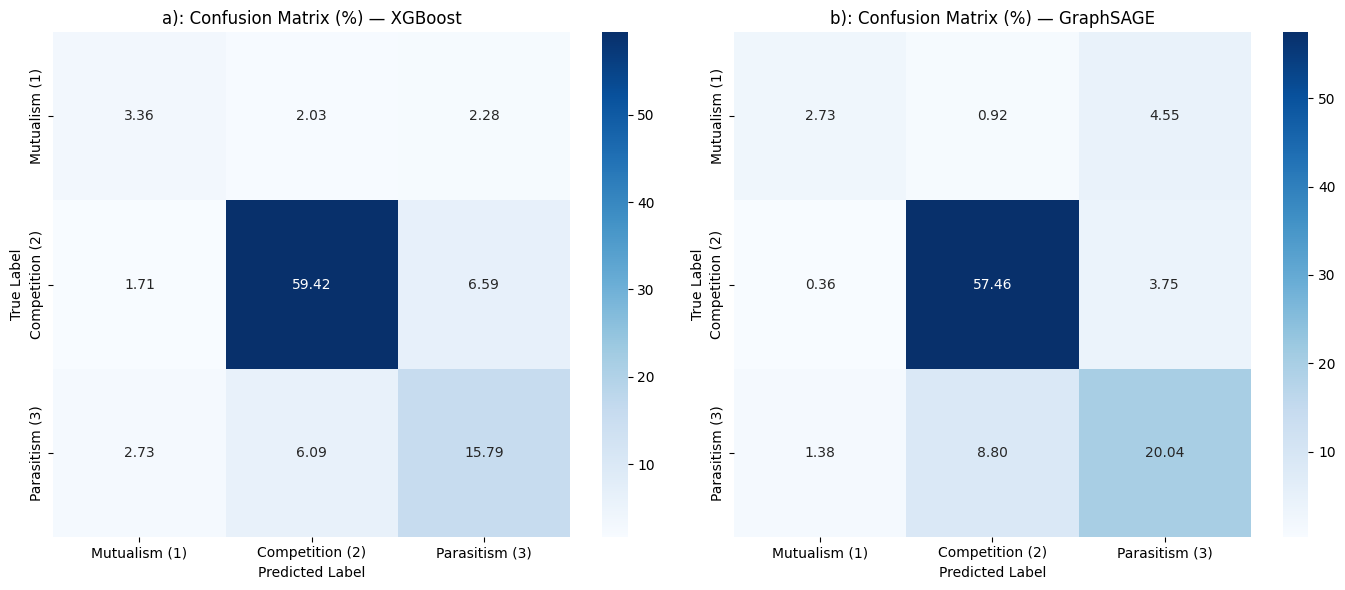}
\begin{center}
\centering
\caption{Comparison of confusion matrices for XGBoost \cite{nestor2023interactions} and GraphSAGE in predicting two-way interaction signs.  }\label{two_way}
\end{center}
\end{figure}

In our study, we observed that  mutualistic interactions (+,+) were particularly challenging to predict accurately, often being misclassified as parasitic interactions (+,-). This challenge in predicting positive one-way effects is consistent with ecological research, which suggests that mutualistic interactions are complex and difficult to model because of their intricate dynamics. Table.\ref{tab:performance_comparison} summarizes the  results for different two-way interactions. Figure.\ref{two_way} shows the comparison of confusion matrix of node classification of two-way interaction classes.  The proposed edge-graph-based models demonstrated good performance in two-way sign prediction, achieving an F1-score of  of 43.1\% and 68.4\% slightly outperforming XGBoost in predicting mutualism and parasitism, while XGBoost is slightly better at predicing competition.
Among the two-way sign predictions, competition (-,-) is the most frequently identified interaction type (57.39\%). In contrast, mutualistic interactions (+,+) are particularly challenging to predict. This difficulty likely arises from the inherent challenge of detecting positive one-way effects, which may cause the model to misclassify them as parasitism (+,-) (20.21\%) or, less frequently, as mutualism itself (3.48\%). These misclassifications may also reflect the imbalance in class distributions, as mutualistic cases are relatively underrepresented in the dataset.

\begin{table}[ht]
\centering
\caption{Classification performance for microbial two-way interactions using GraphSAGE and XGBoost.}
\begin{tabular}{lcccccc}
\hline
\multirow{2}{*}{\textbf{Class}} & \multicolumn{3}{c}{\textbf{GraphSAGE}} & \multicolumn{3}{c}{\textbf{XGBoost}\cite{nestor2023interactions}} \\
\cline{2-7}
 & \textbf{Precision} & \textbf{Recall} & \textbf{F1-score} & \textbf{Precision} & \textbf{Recall} & \textbf{F1-score} \\
\hline
Mutualism    & \textbf{0.611} & 0.333 & \textbf{0.431} & 0.430 & \textbf{0.440} & 0.430 \\
Competition  & 0.855 & \textbf{0.933} & 0.893 &\textbf{ 0.880} & 0.920 & \textbf{0.900} \\
Parasitism   &\textbf{ 0.707} &\textbf{ 0.663} & \textbf{0.684} & 0.640 & 0.570 & 0.600 \\
\hline
\textbf{Accuracy} & \multicolumn{3}{c}{\textbf{0.811}} & \multicolumn{3}{c}{0.780} \\
\hline
\end{tabular}
\label{tab:performance_comparison}
\end{table}

\section{Limitations}

Microbial interactions offer important insights into community dynamics but are challenging to measure. While our results show that microbial interactions can be predicted under controlled laboratory conditions, it is unclear how well this generalizes to natural environments. The model focuses on pairwise interactions and does not capture higher-order effects, which may be influenced by additional species in complex communities. Performance also depends on the availability and quality of input features, such as monoculture growth yield. Moreover, graph neural networks are computationally intensive and require access to HPC resources, potentially limiting reproducibility. We might also need to consider  whether there are better ways to structure the graphs to encode the experimental information more effectively.
These limitations suggest directions for future work, including testing on broader datasets, incorporating higher-order interactions through approaches such as Topological Deep Learning (TDL), improving computational efficiency, and developing lightweight methods for resource-limited settings.

\section{Conclusion}
In this work, we illustrate that an edge‑focused graph learning approach can reveal both the structure and direction of interactions in complex biological systems. By converting microbial networks into edge‑graphs and applying inductive GNNs like GraphSAGE, we capture the properties of interactions directly, rather than only the traits of species. 
We tested this approach with data of bacterial communities, where it  outperforms XGBoost as a strong baseline. The method is generalizable and could be applied to other ecological systems, epidemiological networks, or any network with diverse, directional links. In the long run, by embedding biologically meaningful features at each edge, geometric deep learning models can achieve high predictive accuracy while reflecting real-world interaction patterns, bridging the gap between data-driven prediction and biological understanding.

\bibliographystyle{plainnat}

\appendix

\section{Technical Appendices and Supplementary Material}
\subsection{Features names information}\label{appendix:features}
Table~\ref{tab:features} summarizes the features used for microbial interaction prediction, including monoculture growth measurements, metabolic dissimilarity, and principal components derived from carbon source utilization and phylogenetic embeddings.

\renewcommand{\thetable}{A.\arabic{table}}
\setcounter{table}{0}
\begin{table}[ht]
\centering
\caption{Features used for microbial interaction prediction.}
\begin{tabular}{ll}
\hline
\textbf{Feature} & \textbf{Description} \\
\hline
\multicolumn{2}{l}{\textbf{Monoculture growth features}} \\
monoGrow\_x       & Monoculture growth of species $x$ \\
monoGrow\_y       & Monoculture growth of species $y$ \\
monoGrow24\_x     & 24-hour monoculture growth of species $x$ \\
monoGrow24\_y     & 24-hour monoculture growth of species $y$ \\
\hline
metDis            & Metabolic dissimilarity between species $x$ and $y$ \\
\hline
\multicolumn{2}{l}{\textbf{Carbon source PCA components (90\% variance)}} \\
carbon\_component\_0 & Principal component 0 of metabolic profiles \\
carbon\_component\_1 & Principal component 1 of metabolic profiles \\
carbon\_component\_2 & Principal component 2 of metabolic profiles \\
carbon\_component\_3 & Principal component 3 of metabolic profiles \\
\hline
\multicolumn{2}{l}{\textbf{Phylogenetic PCA components (95\% variance)}} \\
phy\_strain\_component\_0\_x & Principal component 0 of phylogenetic embedding (species $x$) \\
phy\_strain\_component\_1\_x & Principal component 1 of phylogenetic embedding (species $x$) \\
phy\_strain\_component\_0\_y & Principal component 0 of phylogenetic embedding (species $y$) \\
phy\_strain\_component\_1\_y & Principal component 1 of phylogenetic embedding (species $y$) \\
\hline
\end{tabular}
\label{tab:features}
\end{table}

\subsection{Assessment metrics}\label{appendix:metrics}
The evaluation metrics used in this study are defined as follows:

\begin{equation}
\text{Accuracy} = \frac{TP + TN}{TP + TN + FP + FN}
\label{eq:accuracy}
\end{equation}

\begin{equation}
\text{Sensitivity} = \frac{TP}{TP + FN}
\label{eq:sensitivity}
\end{equation}

\begin{equation}
\text{Precision} = \frac{TP}{TP + FP}
\label{eq:precision}
\end{equation}

\begin{equation}
\text{F1} = 2 \times \frac{\text{Precision} \times \text{Recall}}{\text{Precision} + \text{Recall}}
\label{eq:f1}
\end{equation}
Where True Positive (TP) denotes the positive instances correctly identified as positive, True Negative (TN) indicates the negative instances correctly identified as negative, False Positive (FP) represents the negative instances incorrectly identified as positive, and False Negative (FN) refers to the positive instances incorrectly identified as negative.


\end{document}